\documentclass[preprint,number,review,12pt]{elsarticle} 

\usepackage[top=2.5cm,bottom=2.5cm,left=2.5cm,right=2.5cm]{geometry}
\usepackage{amssymb}
\usepackage{xcolor}
\usepackage{graphicx}
\graphicspath{ {./images/} }
\usepackage{multirow} 
\usepackage[centertags]{amsmath} \usepackage{amsfonts}
\usepackage{amsthm}
\usepackage{textcomp}
\usepackage{dblfloatfix}
\usepackage{caption}
\usepackage{cleveref}
\usepackage{subcaption}
\usepackage{lineno}
\DeclareMathAlphabet{\mathpzc}{OT1}{pzc}{m}{it}
\usepackage{amsmath}
\usepackage{bbm}
\usepackage[mathscr]{euscript}
\usepackage{gensymb}
\usepackage{color}
\usepackage{comment}
\usepackage{float}
\usepackage[flushleft]{threeparttable}
\usepackage{booktabs}
\usepackage{amsmath} 
\usepackage{url} 

\newcommand{\Expect}{{\rm I\kern-.3em E}}

\journal{Preprint submitted to Engineering Structures}

\begin{document}

\begin{frontmatter}



\title{Deep Learning-Based Metamodeling of Nonlinear Stochastic Dynamic Systems under Parametric and Predictive Uncertainty}


\author[Umich]{Haimiti Atila}
\ead{hatila@umich.edu}
\author[Umich]{Seymour M.J. Spence\corref{cor1}\fnref{fn1}}
\ead{smjs@umich.edu}
\cortext[cor1]{Corresponding author. The corresponding author is an editor of this journal. In accordance with policy, the corresponding author was blinded to the entire peer review process.}

\address[Umich]{Department of Civil and Environmental Engineering, University of Michigan, Ann Arbor, MI 48109, USA}


\begin{abstract}

Modeling high‐dimensional, nonlinear dynamic structural systems under natural hazards presents formidable computational challenges, especially when simultaneously accounting for uncertainties in external loads and structural parameters. Studies have successfully incorporated uncertainties related to external loads from natural hazards, but few have simultaneously addressed loading and parameter uncertainties within structural systems while accounting for prediction uncertainty of neural networks. To address these gaps, three metamodeling frameworks were formulated, each coupling a feature‐extraction module—implemented through a multi‐layer perceptron (MLP), a message‐passing neural network (MPNN), or an autoencoder (AE)—with a long short‐term memory (LSTM) network using Monte Carlo dropout and a negative log-likelihood loss. The resulting architectures (MLP‐LSTM, MPNN‐LSTM, and AE‐LSTM) were validated on two case studies: a multi‐degree‐of‐freedom Bouc–Wen system and a 37-story fiber‐discretized nonlinear steel moment‐resisting frame, both subjected to stochastic seismic excitation and structural parameter uncertainty. All three approaches achieved low prediction errors: the MLP‐LSTM yielded the most accurate results for the lower‐dimensional Bouc–Wen system, whereas the MPNN‐LSTM and AE‐LSTM provided superior performance on the more complex steel‐frame model. Moreover, a consistent correlation between predictive variance and actual error confirms the suitability of these frameworks for active‐learning strategies and for assessing model confidence in structural response predictions.

\end{abstract}



\end{frontmatter}


\section{Introduction}
\label{Sec:Intro}

Nonlinear dynamic response analysis is frequently required for nonlinear multi-degree-of-freedom (MDOF) structural systems subjected to time-varying excitations. Such analyses capture essential dynamic behaviors and allow for the calculation of key performance metrics, including peak responses, residual responses, and cumulative damage assessments. Typically, dynamic response analyses are conducted numerically using various integration schemes. However, for high-dimensional structural systems exhibiting complex nonlinear behavior, these numerical methods become computationally intensive. This challenge intensifies further in applications like uncertainty quantification and design optimization, which require multiple simulations under varying loading sequences and structural model parameters.

Metamodeling techniques are essential for reducing such computational barriers. Researchers have previously employed approaches such as kriging, radial basis functions, support vector machines, and polynomial regression to approximate the input-output relationship of high-fidelity numerical models, significantly reducing computational demands \cite[e.g.][]{KROETZ2017, Kianifar2020, TEIXEIRA2021, NEGRIN2023}. Nonetheless, traditional metamodeling approaches frequently struggle to accurately replicate dynamic response time histories, particularly under stochastic excitations. These types of problems, defined as sequence-to-sequence learning tasks, can become prohibitively high-dimensional due to the necessity of predicting responses across numerous discrete time steps. Many works have used methodologies such as Multi-Input Multi-Output Nonlinear AutoRegressive eXogenous (MIMO-NARX) models to overcome such a challenge; however, NARX models are limited by the need for predefined function forms, which generally require prior
knowledge of the system for identification \cite{spiridonakos2015metamodeling,mai2016surrogate,chuang2019rapid, Li21}. More recently, to break free from the limitation of the NARX models, researchers have turned their attention to deep learning frameworks, such as recurrent neural networks, physics-informed machine learning architectures, and neural operators to improve metamodeling capabilities for dynamic nonlinear structural systems  \cite{ning2023lstm,zhang2019deep,wu2019deep,simpson2021machine,li2022metamodeling,Li_2024,liao2023attention,Atila2025PhysicsInformedLSTM,goswami2025neural,zhang2020physics}. 

Despite the demonstrated capability of recent machine learning methods to reproduce dynamic responses under stochastic excitation and generalize across loading sequences, many existing studies fail to account for two critical sources of uncertainty: (1) structural parameter uncertainty—encompassing variations in mass, damping, material properties, and other physical characteristics—and (2) prediction uncertainty, which includes both epistemic uncertainty arising from limited training data and aleatoric uncertainty stemming from inherent variability or modeling limitations in the learning algorithm. Neglecting structural variability undermines the generalizability of these approaches when applied to structures with differing physical attributes. While, disregarding prediction uncertainty hinders the reliable interpretation of metamodel outputs and limits the model's utility for informed decision-making during the model's inference time.

Although several recent studies have begun to incorporate structural parameter variability under stochastic loading (e.g., \cite{zhou2025parameter,kundu2022long,zhong2025multi,ning2025surrogate,zhang2024rlstm}), and in one case have attempted to quantify predictive uncertainty (e.g., \cite{ning2025surrogate}), such efforts remain limited. In particular, the latter addresses only aleatoric uncertainty while neglecting epistemic contributions. Furthermore, the predictive scope in these works is typically restricted to a subset of degrees of freedom rather than encompassing the full structural system. To the best of the authors’ knowledge, no existing study has developed a comprehensive metamodel capable of predicting the complete time-history response across all degrees of freedom in large-scale nonlinear structural systems, while simultaneously accounting for stochastic excitations, structural parameter uncertainty, and both epistemic and aleatoric forms of uncertainty.

To address the aforementioned research gap, this study introduces three metamodeling frameworks that not only generalize across diverse stochastic excitation scenarios and structural parameter variations but also provide prediction uncertainty quantification. Each framework consists of two interdependent modules: a feature extraction/fusion module and a time-series prediction module. The feature extraction/fusion module is designed to distill salient information from both the structural configuration and stochastic excitation into compact feature vectors. Three distinct architectures are explored for this purpose: a multi-layer perceptron (MLP), for its architectural simplicity; a message passing neural network (MPNN), which captures relational dependencies by leveraging the system's graph-based representation; and an autoencoder (AE), which facilitates nonlinear dimensionality reduction, particularly advantageous for high-dimensional systems.

These fused feature vectors are subsequently processed by a long short-term memory (LSTM) network integrated with Monte-Carlo dropout and trained using the negative log-likelihood loss function. This configuration enables the LSTM to learn temporal dynamics while simultaneously quantifying both aleatoric and epistemic uncertainty. To mitigate the computational burden and enhance training efficiency, wavelet-based approximations are incorporated into the LSTM module.

The resulting architectures—MLP–LSTM, MPNN–LSTM, and AE–LSTM—are evaluated using two case studies: (1) a multi-degree-of-freedom Bouc–Wen system, and (2) a 37-story nonlinear fiber-discretized steel moment-resisting frame. Both systems are subjected to stochastic seismic excitations and incorporate uncertainty in their structural properties. These case studies are designed to assess and compare the performance of the proposed methods, with the goal of elucidating their respective strengths and limitations in capturing the dynamic response and associated predictive uncertainties. The MLP–LSTM model, owing to its architectural simplicity, was designated as the baseline against which the more advanced MPNN–LSTM and AE–LSTM variants were benchmarked.

The remainder of this paper is organized as follows. Sec. \ref{Sec:Problem Setting} provides a formal statement of the problem. Sec.~\ref{Sec:Proposed Approaches} presents the detailed architectures of the three proposed schemes. Sec. ~\ref{Sec:Case Study 1} and \ref{Sec:Case Study 2} describe the experimental setups and results for the Bouc–Wen system and the fiber-discretized nonlinear steel moment-resisting frame system, respectively. Finally, Sec.~\ref{Sec:Conclusion} offers concluding remarks and discusses the limitations and potential extensions of the current work.
%


\section{Problem Setting}
\label{Sec:Problem Setting}
The dynamic response of nonlinear structural systems subjected to stochastic excitation can be characterized by solving the set of nonlinear ordinary differential equations:
\begin{equation}\label{eq:EOM}
\mathbf{M(\mathbf{\Gamma})} \ddot{\mathbf{u}}(t)+\mathbf{C( \mathbf{\Gamma} )} \dot{\mathbf{u}}(t)+\mathbf{f}_{nl}(t;\mathbf{u}(t),\mathbf{\Gamma})= \mathbf{F}(t)
\end{equation}
where $\mathbf{u}(t)$, $\dot{\mathbf{u}}(t)$, and $\ddot{\mathbf{u}}(t)$ denote the displacement, velocity, and acceleration vectors, respectively, for a system with $n$ degrees of freedom; the matrices $\mathbf{M}(\mathbf{\Gamma})$ and $\mathbf{C}(\mathbf{\Gamma})$ are the mass and damping matrices of size $n \times n$, both of which depend on the random vector $\mathbf{\Gamma}$ that encapsulates uncertainty in structural parameters; $\mathbf{f}_{\mathrm{nl}}(t; \mathbf{u}(t), \mathbf{\Gamma})$ represents the nonlinear restoring force, which is generally a function of the displacement and the uncertain parameters; and $\mathbf{F}(t)$ corresponds to the external stochastic excitation vector applied to the structure.

To numerically integrate Eq.~(\ref{eq:EOM}), a time discretization scheme is usually employed, resulting in the form:
\begin{equation}\label{eq:EOM_discret}
\mathbf{M(\mathbf{\Gamma})} \ddot{\mathbf{u}}_i+\mathbf{C( \mathbf{\Gamma} )} \dot{\mathbf{u}}_i+\mathbf{f}_{nl}(\mathbf{u}_i,\mathbf{\Gamma})= \mathbf{F}_i
\end{equation}
where $\mathbf{u}_i$, $\dot{\mathbf{u}}_i$, and $\ddot{\mathbf{u}}_i$ denote the displacement, velocity, and acceleration vectors with size $n\times1$ at the $i$-th time step, while $\mathbf{F}_i$ is the corresponding external excitation. The nonlinear force $\mathbf{f}_{nl}(\mathbf{u}_i,\mathbf{\Gamma})$ is evaluated at each time step based on the current displacement and pre-defined structural parameters. Time integration proceeds incrementally with a time step $\Delta t$, generating a sequence of response vectors ${\mathbf{u}_1, \mathbf{u}_2, \ldots, \mathbf{u}_T}$ in response to the sequence of external forces ${\mathbf{F}_1, \mathbf{F}_2, \ldots, \mathbf{F}_T}$, where $T$ denotes the total number of time steps in the simulation.

One of the primary objectives of this study is to develop neural network–based metamodels, denoted by $\phi_{\theta}(\cdot)$, capable of accurately predicting the full time-history response sequence of the structural system governed by Eq.~\eqref{eq:EOM_discret}. These metamodels take as input the sequence of external stochastic excitations and the corresponding structural parameters and output the predicted dynamic response over time (i.e., $\hat{\mathbf{u}}_1, \hat{\mathbf{u}}_2, \ldots, \hat{\mathbf{u}}_T$) with $\theta$ representing the set of trainable parameters of the neural network. 

Despite the expressive power of neural network-based metamodels, their predictions are inherently accompanied by uncertainty, which arises from both the limited availability of training data and the structural choices in model architecture. Following the framework introduced by \citet{gal2017concrete}, such prediction uncertainty can be decomposed into two components: epistemic uncertainty and aleatoric uncertainty. Epistemic uncertainty, often referred to as reducible uncertainty, originates from the finite training data and diminishes as the size of the training data increases. In contrast, aleatoric uncertainty—commonly termed irreducible uncertainty—persists even with an infinite dataset.

In the context of machine learning, epistemic uncertainty reflects the variability in model parameters $\theta$ induced by computational constraints and limited training samples. Aleatoric uncertainty, by comparison, encapsulates the intrinsic expressive limitations of the model architecture in accurately capturing the true input–output relationship. These limitations may result from noisy data, a restricted number of trainable parameters, or suboptimal convergence behaviors associated with the employed learning algorithms \cite{schweighofer2024information,helton1997uncertainty}. When interpreted through the lens of the bias–variance decomposition—a classic concept in machine learning—epistemic uncertainty aligns with the variance component, whereas aleatoric uncertainty encompasses both the bias introduced by model mis-specification and the noise in the observations \cite{apostolakis1990concept,bengs2022pitfalls,kato2022view}. Since the present work assumes the input-output data to be noise-free, the primary source of aleatoric uncertainty is the model mis-specification.
A further aim of this work is therefore to devise and evaluate methods for quantifying both epistemic and aleatoric uncertainty in the predictions of $\phi_{\theta}(\cdot)$. 
%


\section{The Proposed Approaches}
\label{Sec:Proposed Approaches}
In this section, the specific configuration of the wavelet transform used in this study is introduced in Sec.~\ref{Sec:Wavelet_Transform}. Subsequently, Sec.~\ref{Sec:Feature_Module} details the architecture of the feature extraction and fusion modules, including the multi-layer perceptron (Sec.~\ref{Sec:MLP}), message-passing neural network (Sec.~\ref{Sec:MPNN}), and autoencoder (Sec.~\ref{Sec:AE}). Each of these sub-sections concludes with a clear specification of the final output generated by these models. Finally, Sec.~\ref{Sec:Time-series} outlines the implementation of the long short-term memory (LSTM) network, Monte-Carlo dropout, and the negative log-likelihood loss function for predicting structural time-history responses with quantified uncertainty, using the outputs of the feature extraction/fusion module as input.


\subsection{Data-Processing: Wavelet Transformation}
\label{Sec:Wavelet_Transform}

The input/output sequences of Eq.~(\ref{eq:EOM_discret}) typically encompass a large number of time steps. Consequently, the LSTM layers require a large number of cells, which may result in a considerable computational burden and memory overhead. To mitigate this, rather than directly predicting the structural response from the external excitation force in the original high-dimensional temporal space, these inputs and outputs are transformed into the wavelet domain. This serves as a data pre-processing technique that significantly reduces the temporal dimension of the data. 

Within this setting, each component of a discrete representation of a vector-valued stochastic process, i.e., each component of $\mathbf{F}_i$, $\mathbf{u}_i$, can be written in the form:    
\begin{align}\label{eq:Wavelet_ex}
\chi_i = \sum_{k} a_{J,k} \Omega_{J,k}(i) + \sum_{j=0}^{J-1} \sum_{k} d_{j,k} \psi_{j,k}(i)
\end{align} 
where $J$ denotes the coarsest decomposition level; $\Omega_{J,k}(i)$ and $a_{J,k}$ are the scaling functions and their associated approximation coefficients at level $J$, while $\psi_{j,k}(i)$ and $d_{j,k}$ are the wavelet functions and detail coefficients at level $j$. Dyadic wavelets are employed in this work owing to their demonstrated efficacy in related metamodeling applications \cite{le2015reduced,wang2020knowledge,li2022metamodeling,Li_2024}. In particular, an effective approximation of Eq. (\ref{eq:Wavelet_ex}) can be defined by considering only the contribution of the scaling and approximation coefficients at level $J$. Each realization of the vector‐valued stochastic process is characterized by its level-$J$ wavelet approximation coefficients, $a_{J,k}$. Consequently, the LSTM network is trained to learn the mapping between sequences of input and output approximation coefficients rather than the full time-domain signals. Because the coefficient sequences are substantially shorter than their corresponding time-domain series, the number of recurrent cells per layer is markedly reduced. This compression both accelerates network training and often yields more accurate predictions. Once the LSTM has generated the predicted coefficient sequence, the time-domain response is recovered by applying the inverse wavelet transform (Eq.~\ref{eq:Wavelet_ex}), retaining only the approximation terms up to level $J$.

In this framework, the transformed forcing vector at a particular $k$ is denoted:  
\[
\mathbf{a}_{J,k}^{(\mathbf{F})}
= \bigl\{a_{J,k}^{(\mathbf{F}^{(1)})},\,a_{J,k}^{(\mathbf{F}^{(2)})},\,\ldots,\,a_{J,k}^{(\mathbf{F}^{(n)})}\bigr\}
\]  
and the corresponding response vector is:  
\[
\mathbf{a}_{J,k}^{(\mathbf{u})}
= \bigl\{a_{J,k}^{(\mathbf{u}^{(1)})},\,a_{J,k}^{(\mathbf{u}^{(2)})},\,\ldots,\,a_{J,k}^{(\mathbf{u}^{(n)})}\bigr\}
\]  
Training the LSTM on these compact coefficient representations thus enables efficient sequence modeling. 


\subsection{Feature Extraction/Fusion Module}
\label{Sec:Feature_Module}
\subsubsection{Multi-layer perceptron}
\label{Sec:MLP}

Multi‐layer perceptrons (MLPs) are widely employed for fusing different feature representations into a single, informative embedding \cite{tolstikhin2021mlp}. Formally, an $L$‐layer MLP is defined as:
\begin{equation}\label{eq:MLP}
\hat{\mathbf{y}}^{\mathrm{mlp}} \;=\;\rho^{(L)}\Bigl(\mathbf{W}^{(L)}\,\rho^{(L-1)}\bigl(\dots\,\rho^{(1)}(\mathbf{W}^{(1)}\mathbf{x}^{\mathrm{mlp}}+\mathbf{b}^{(1)})\dots\bigr)+\mathbf{b}^{(L)}\Bigr)
\end{equation}
where $\mathbf{x}$ denotes the input vector—often formed by concatenating multiple feature vectors—$\mathbf{W}^{(l)}$ and $\mathbf{b}^{(l)}$ are the trainable weight matrix and bias vector of the $l$-th layer, and $\rho^{l}(\cdot)$ is that layer’s nonlinear activation function. Optimal weight matrices $\{\mathbf{W}^{(l)}\}_{l=1}^L$ and bias vectors $\{\mathbf{b}^{(l)}\}_{l=1}^L$, which yield an output $\hat{\mathbf{y}}^\mathrm{mlp}$ that efficiently fuses information from the input vector $\mathbf{x}^{\mathrm{mlp}}$, are obtained by end-to-end training of the MLP jointly with downstream neural network modules that consume $\hat{\mathbf{y}}^\mathrm{mlp}$ for task-specific prediction. 

In the present study, the input feature vector $\mathbf{x}^{\mathrm{mlp}}$ for the MLP is constructed by concatenating the wavelet-transformed stochastic load vector with the structural random parameter vector $\mathbf{\Gamma}$. Since $\mathbf{\Gamma}$ directly influences the mass matrix, the damping matrix, and the restoring‐force behavior, it can be regarded as a summary of the structure’s characteristics. At particular wavelet index $k$, the combined input vector $\mathbf{x}_{k}^\mathrm{mlp}$ is therefore given by:
\begin{equation}\label{eq:X_concat_mlp}
\mathbf{x}_{k}^{\mathrm{mlp}} \;=\; 
\begin{Bmatrix}
\mathbf{a}_{J,k}^{(\mathbf{F})} \\
\mathbf{\Gamma}
\end{Bmatrix}
\end{equation}
Subsequently, the output vector of the MLP, $\hat{\mathbf{y}}^{\mathrm{mlp}}_k$, is passed to the time‐series prediction module at each step $k$ to forecast the system’s response, $\mathbf{a}_{J,k}^{(\mathbf{u})}$, at the corresponding instant. Consequently, the MLP and the time‐series prediction module are trained jointly in an end‐to‐end manner.


\subsubsection{Message passing neural network}
\label{Sec:MPNN}

The message passing neural network (MPNN) framework, which unifies and generalizes many earlier graph-based models, has been widely recognized for its effectiveness in processing and providing abstract representations of graph-structured data through capturing relational interaction within the graph \cite{gilmer2020message,ying2018graph}. A graph ($G$) is defined as $G = <V,E>$, where $V$ is a set of nodes and $E$ is a set of edges connecting the nodes. In message-passing neural networks, each node is assumed to have a node feature describing the node, and an optional edge feature describing the property of the edge. Thus, $
V = \{ \mathbf{v}_{p} \}_{p=1}^{N^v}$, where $\mathbf{v}_{p}$ is the node feature vector of node $p$, and  $
E = \{ \mathbf{e}_{q} \}_{q=1}^{N^e}$, where $\mathbf{e}_{q}$ is the optional edge feature vector of edge $q$. Graph‐based representations of structural systems are commonly formulated in two ways. One approach models structural joints as nodes and members (beams and columns) as edges, while the alternative represents members as nodes interconnected by edges corresponding to the joints that link them \cite{cao2024vertex,chang2020learning,chou2024structgnn,kuo2024gnn,liu2025graph}. In this work, the latter formulation was adopted because the random variables in $\mathbf{\Gamma}$ generally pertain to structural members. By encoding members as nodes, the need for explicit edge features is eliminated, thereby reducing computational overhead. Consequently, each node’s feature vector encapsulates both the inherent properties of its corresponding member and any global structural parameters—such as damping ratios that are applied uniformly across the system. Similar to prior studies \cite{chang2020learning,chou2024structgnn}, the current study also adds a virtual node linked to all member‐nodes to facilitate global information aggregation during subsequent message‐passing operations.

Message-passing enables nodes to iteratively exchange information with their neighbors and update their representations based on local relational context. In this work, the message-passing of the feature of node \(p\) at message-passing iteration \(m+1\) is computed as:
\begin{equation}\label{eq:MPNN_EQ}
\mathbf{v}_p^{(m+1)}
= g\Bigl(
    \mathbf{v}_p^{(m)} + \mathbf{v}_p^{(m-1)},\;
    \bigoplus_{j \in \mathcal{N}(p)}
        f\bigl(
            \mathbf{v}_p^{(m)} + \mathbf{v}_p^{(m-1)},\;
            \mathbf{v}_j^{(m)} + \mathbf{v}_j^{(m-1)}
        \bigr)
\Bigr)
\end{equation}
where \(\mathbf{v}_p^{(m)}\) denotes the embedding of node \(p\) after \(m\) message‐passing iterations, \(\mathcal{N}(p)\) is the set of its neighbors, and \(\bigoplus\) indicates an averaging aggregation. The functions \(f(\cdot)\) and \(g(\cdot)\) are each realized by a feedforward neural network. Intuitively, with Eq. (\ref{eq:MPNN_EQ}), at each iteration, nodes compute messages through a learnable function of their own and neighboring node features, then aggregate these messages through permutation-invariant operations such as averaging, and apply a separate update function to yield refined node states. By repeating this process for $M$ steps, information propagates across the graph up to $M$ hops, allowing node embeddings to capture increasingly global structural and feature interactions. Furthermore, to mitigate the tendency towards over-smoothing of node feature vectors across successive message‐passing layers, a residual connection is added.

Upon completion of \(M\) message‐passing steps, a global mean pooling operation over all \(N_v\) nodes is used to yield a compact graph‐level representation:
\begin{equation}\label{eq:Mean_Pool_refined}
\hat{\mathbf{\Gamma}} = \frac{1}{N_v} \sum_{p=1}^{N_v} \mathbf{v}_p^{(M)} \,.
\end{equation}
The pooled vector $\hat{\boldsymbol{\Gamma}}$ is assumed to capture the key characteristics of the underlying structural graph, and hence of the entire structural system. Consequently, at each wavelet scale index $k$, the excitation-driven coefficient vector $\mathbf{a}_{J,k}^{(\mathbf{F})}$ is concatenated with $\hat{\boldsymbol{\Gamma}}$ and passed through an auxiliary feedforward neural network parameterized by $\theta_{\mathrm{fnn}}$. The output of this network then serves as the feature input to the subsequent time-series prediction module, which forecasts the system’s response at the corresponding instant. Formally, the MPNN’s output feature vector at index $k$ is given by:
\begin{equation}\label{eq:X_concat_mpnn}
\mathbf{x}_{k}^{\mathrm{mpnn}}
= 
\mathrm{FNN}_{\theta_{\mathrm{fnn}}}\!\bigl(\{\mathbf{a}_{J,k}^{(\mathbf{F})},\,\hat{\boldsymbol{\Gamma}}\bigr\})
\end{equation}
Similar to the approach presented in Sec \ref{Sec:MLP}, the message-passing neural network and the time-series prediction module are trained jointly in an end-to-end manner to predict $\mathbf{a}_{J,k}^{(\mathbf{u})}$ given $\mathbf{x}_{k}^{\mathrm{mpnn}}$.

\subsubsection{Autoencoder}
\label{Sec:AE}

While the approaches presented in Sec.~\ref{Sec:MLP} and Sec.~\ref{Sec:MPNN} aims to direct forecasting of the $n$-dimensional response vector $\mathbf{a}_{J,k}^{(\mathbf{u})}$, projecting these high-dimensional outputs onto a lower-dimensional space has been shown to markedly enhance both computational efficiency and predictive accuracy \cite{li2022metamodeling,Li_2024,kontolati2024learning,simpson2021machine}. Accordingly, this study implements an autoencoder as a third strategy to reduce the dimensionality of $\mathbf{a}_{J,k}^{(\mathbf{u})}$ prior to prediction, such that the subsequent time-series prediction module needs only to learn the mapping from the stochastic input load-related wavelet coefficients to the corresponding low-dimensional latent representation of $\mathbf{a}_{J,k}^{(\mathbf{u})}$. The autoencoders—by virtue of their capacity to learn nonlinear mappings—offer superior reduction performance compared with linear techniques such as principal component analysis (PCA) \cite{hinton2006reducing}. In an autoencoder framework, an encoder transforms the original high-dimensional input into a compact latent representation, while a decoder reconstructs the input from that representation.

In the present study, the encoder $E_{\theta_e}$ and decoder $D_{\theta_d}$ are defined by feedforward neural network as:
\begin{equation}\label{eq:MLP_Enco}
\mathbf{r} \;=\; E_{\theta_{e}}(\mathbf{x}^{(e)})
\;=\;
\mathbf{W}_e^{(L_e)}\,
\rho_e^{(L_e-1)}\Bigl(\dots\rho_e^{(1)}\bigl(\mathbf{W}_e^{(1)}\mathbf{x}^{(e)} + \mathbf{b}_e^{(1)}\bigr)\dots\Bigr)
+ \mathbf{b}_e^{(L_e)}
\end{equation}
\begin{equation}\label{eq:MLP_Deco}
\hat{\mathbf{y}}^{(d)}
\;=\;
D_{\theta_d}(\mathbf{x}^{(d)})
\;=\;
\mathbf{W}_d^{(L_d)}\,
\rho_d^{(L_d-1)}\Bigl(\dots\rho_d^{(1)}\bigl(\mathbf{W}_d^{(1)}\mathbf{x}^{(d)} + \mathbf{b}_d^{(1)}\bigr)\dots\Bigr)
+ \mathbf{b}_d^{(L_d)}
\end{equation}
where
\[
\theta_e = \bigl\{\mathbf{W}_e^{(\ell)},\,\mathbf{b}_e^{(\ell)}\bigr\}_{\ell=1}^{L_e},
\qquad
\theta_d = \bigl\{\mathbf{W}_d^{(\ell)},\,\mathbf{b}_d^{(\ell)}\bigr\}_{\ell=1}^{L_d}
\]
and $\rho_e^{(\cdot)}$, $\rho_d^{(\cdot)}$ denote element-wise activation functions.  Because the structural response is influenced by the structural parameters, both encoder and decoder inputs are augmented with $\mathbf{\Gamma}$. At wavelet index $k$, the network inputs and outputs are:
\begin{equation}\label{eq:Xe_concat}
\mathbf{x}^{(e)}_{k} =
\begin{bmatrix}
\mathbf{a}_{J,k}^{(\mathbf{F})} \\[4pt]
\mathbf{\Gamma}
\end{bmatrix}, 
\quad
\mathbf{x}^{(d)}_{k} =
\begin{bmatrix}
\mathbf{r}_k \\[4pt]
\mathbf{\Gamma}
\end{bmatrix}
\end{equation}
where $\mathbf{r}_k = E_{\theta_e}(\mathbf{x}^{(e)}_{k})$ and $\hat{\mathbf{y}}^{(d)}_{k} = D_{\theta_d}(\mathbf{x}^{(d)}_{k})$.  $\mathbf{r}_k$ is a vector of size $n_r \times 1$, where $n_r \ll n$. Training proceeds by minimizing the reconstruction error between $\hat{\mathbf{y}}^{(d)}_{k}$ and the true response $\mathbf{a}_{J,k}^{(\mathbf{u})}$. The autoencoder is trained separately from the time-series module. Once the low-dimensional representation, ${\mathbf{r}}_{k}$, is obtained, the remaining procedure follows that described in Sec.~\ref{Sec:MLP}; however, the target data is changed from $\mathbf{a}_{J,k}^{(\mathbf{u})}$ to ${\mathbf{r}}_{k}$ for the subsquent time-series prediction module.

\subsection{Time Series Prediction Module}
\label{Sec:Time-series}

\subsubsection{Stack-LSTM Architecture}
\label{Sec:Stacked-LSTM}
Due to their ability to incorporate past output information when making current predictions while mitigating gradient explosion or vanishing,  long short-term memory (LSTM) networks emerge as a potent tool for sequence-to-sequence mapping problems \cite{hochreiter1997long,gers2000learning,zhao2017lstm,li2022metamodeling, Li_2024}. In the context of structural dynamics, it is crucial to recognize that the current behavior of a structure is influenced not only by the current external excitation but also by the system's previous response. Therefore, the current work utilizes LSTM networks to learn the dynamics of the structure, leveraging output from the feature extraction/fusion module. To enhance the expressive power of the LSTM, this work employs a stacked LSTM architecture, where the hidden state is passed through multiple layers of LSTMs. Namely, the hidden state of the $l$-th layer of the LSTM at wavelet index $k$ is defined as follows: 
\begin{equation}\label{eq:layered_LSTM}
(\mathbf{h}_{k}^{(l)},\,\mathbf{c}_{k}^{(l)}) 
= \mathrm{LSTM}_{\theta^{(l)}_\mathrm{LSTM}}\!\bigl(\mathbf{h}_{k}^{(l-1)},\,\mathbf{h }_{k-1}^{(l)},\,\mathbf{c}_{k-1}^{(l)}\bigr),
\end{equation}
where $\mathbf{h}_{k}^{l}$ is the hidden state vector of $l$-th layer of the LSTM at wavelet index $k$; $\mathbf{c}_{k}^{l}$ is the cell state vector of $l$-th layer of the LSTM at wavelet index $k$; ${\theta^{(l)}}$ is the parameter of the $l$-th layer of the LSTM. Note, $\mathbf{h}_{k}^{(0)}$ is the input to the LSTM from the feature extraction/fusion module. Collectively, $\theta_{\mathrm{LSTM}} = (\theta^{(1)}_\mathrm{LSTM}, \dots, \theta^{(L)}_\mathrm{LSTM})$ represents the parameters of the $L$-layered LSTM. 

\subsubsection{Quantification of Epistemic Uncertainty through Monte-Carlo Dropout}
\label{Sec:Epistemic_Uncertainty}

Given a finite training dataset \(D_{\mathrm{train}}\), there exist multiple parameter configurations $\theta$ that achieve equivalent fits to $D_{\mathrm{train}}$. Epistemic uncertainty seeks to quantify the dispersion of \(\theta\) under the Bayesian posterior \(P(\theta \mid D_{\mathrm{train}})\). One of the most common approaches for capturing this uncertainty is Bayesian inference implemented through dropout regularization.

Originally introduced by \citet{srivastava2014dropout}, dropout is a regularization technique applied exclusively during the training phase, wherein individual neuron activations are randomly set to zero with a fixed probability prior to being propagated to the subsequent layer. This stochastic masking discourages the co-adaptation of neurons and thereby helps to reduce overfitting. Subsequently, it was demonstrated that retaining dropout during the inference phase enables the model to produce unbiased Monte Carlo estimates of both the predictive mean and variance, under the framework of Bayesian neural networks trained through variational inference \cite{gal2016dropout, gal2016theoretically}. This approach, also known as Monte Carlo dropout (MC-dropout), can be theoretically interpreted as an approximate Bayesian inference method and can also be viewed as a form of model ensembling through stochastic forward passes.

To incorporate MC‐dropout for estimating epistemic uncertainty, the standard layered LSTM update in Eq.~(\ref{eq:layered_LSTM}) is augmented by injecting a dropout mask into the inputs of all intermediate layers. In particular, for layers \(l=2,\dots,L-1\), the hidden–cell update becomes:
\begin{equation}\label{eq:Dropout_LSTM}
(\mathbf{h}_{k}^{(l)},\,\mathbf{c}_{k}^{(l)}) 
= \mathrm{LSTM}_{\theta^{(l)}_{\mathrm{LSTM}}}\!\bigl(\mathbf{z}^{(l-1)} \odot \mathbf{h}_{k}^{(l-1)},\,\mathbf{h}_{k-1}^{(l)},\,\mathbf{c}_{k-1}^{(l)}\bigr),
\end{equation}
where \(\mathbf{z}^{(l-1)}\in\{0,1\}^{d_{l-1}}\) denotes a dropout mask with concrete distribution for the \((l-1)\)-th layer, sampled once per forward pass and held constant across all temporal dimension to preserve the temporal coherence of the recurrent dynamics. The operator \(\odot\) indicates element-wise multiplication. In the modified LSTM architecture, dropout masks remain active at inference time, so each forward pass produces a distinct output. By performing multiple stochastic passes, the network effectively executes Monte Carlo sampling of the posterior over functions, thereby quantifying epistemic uncertainty. Moreover, the dropout probability itself is treated as a learnable parameter and is optimized jointly with the network weights during training.\cite{gal2017concrete}.

\subsubsection{Quantification of Aleatoric Uncertainty through Negative Log-Likelihood Loss Function}
\label{Sec:Aleatoric_Uncertainty}

As noted at the end of Section \ref{Sec:Problem Setting}, this study assumes that the input–output data are noise-free and attributes aleatoric uncertainty primarily to model mis-specification. In this context, mis-specification plays a role parallel to the bias term in the classical bias–variance decomposition, quantifying the discrepancy between the ideal model—i.e., the one that would perfectly capture the true data‐generating process—and the restricted model class actually employed. Thus, depending on the input to the model (i.e., the LSTM model in the present case study), the distribution of the discrepancy also varies. One effective strategy for capturing input-dependent aleatoric uncertainty involves adopting a probabilistic formulation in which the network is trained by maximizing the likelihood of the observed outputs given the inputs, or equivalently by minimizing the negative log-likelihood. In the present study, aleatoric uncertainty is modeled as a Gaussian distribution, a widely used choice owing to its simplicity and effectiveness. Under this assumption, the negative log-likelihood loss for a single training pair $(\mathbf{s}_k,\mathbf{o}_k)$ at wavelet index $k$ and output component $n_o$ can be expressed as:
\begin{equation}\label{eq:nll_loss}
\mathcal{L}_{\mathrm{NLL},k}^{(n_o)}
\bigl(\mathbf{o}_k^{(n_o)} \mid \mu^{(n_o)}(\mathbf{s}_k),\,\sigma^{2^{(n_o)}}(\mathbf{s}_k)\bigr)
= \frac{\bigl(\mathbf{o}_k^{(n_o)} - \mu^{(n_o)}(\mathbf{s}_k)\bigr)^2}
{2\,\sigma^{2^{(n_o)}}(\mathbf{s}_k)} + \frac{1}{2}\,\log\!\bigl(\sigma^{2^{(n_o)}}(\mathbf{s}_k)\bigr).
\end{equation}
where $\mu^{(n_o)}(\mathbf{s}_k)$ and $\sigma^{2^{(n_o)}}(\mathbf{s}_k)$ denote, respectively, the mean and variance predictions produced by the LSTM model for the $n_o$th component of $\mathbf{o}_k$. As observed from Eq.~(\ref{eq:nll_loss}), each component $n_o$ is modeled by an independent Gaussian distribution. The predictive mean $\mu^{(n_o)}(\mathbf{s}_k)$ corresponds to the LSTM’s estimate of $\mathbf{o}_k^{(n_o)}$, whereas $\sigma^{2^{(n_o)}}(\mathbf{s}_k)$ quantifies the aleatoric uncertainty associated with that estimate. Intuitively, the residual term $\frac{\bigl(\mathbf{o}_k^{(n_o)} - \mu^{(n_o)}(\mathbf{s}_k)\bigr)^2}{2\,\sigma^{2^{(n_o)}}(\mathbf{s}_k)}$ forces the network to the increase variance in regions with large errors, so that large errors are ``explained away'' by higher uncertainty.  The regularization term $\frac{1}{2}\,\log\!\bigl(\sigma^{2^{(n_o)}}(\mathbf{s}_k)\bigr)$ penalizes trivial inflation of variance, preventing the model from simply assigning large variances everywhere. For the MLP-LSTM and MPNN-LSTM architectures, set $(\mathbf{s}_k,\mathbf{o}_k) =\bigl(\hat{\mathbf{y}}^{\mathrm{mlp}}_k,\mathbf{a}_{J,k}^{(\mathbf{u})}\bigr)\quad\text{and}\quad
(\mathbf{s}_k,\mathbf{o}_k) =\bigl(\hat{\mathbf{y}}^{\mathrm{mpnn}}_k,\mathbf{a}_{J,k}^{(\mathbf{u})}\bigr)$ respectively. For the AE-LSTM architecture, define $(\mathbf{s}_k,\mathbf{o}_k) =\bigl(\hat{\mathbf{y}}^{\mathrm{mlp}}_k,{\mathbf{r}}_{k}\bigr)$. In order to improve the optimization performance of the negative log‐likelihood loss defined in Eq.~(\ref{eq:nll_loss}), the $\beta$-negative log‐likelihood loss function, a variant of the standard NLL loss that introduces a weighting factor $\beta$ on the log‐variance term, is employed with the default value for $\beta = 0.5$ \cite{seitzer2022pitfalls}. 
\subsubsection{Quantification of Total Prediction Uncertainty}
The predictive uncertainty associated with the $n_o$-th component of the LSTM output $\mathbf{o}_k$ is decomposed into aleatoric and epistemic contributions. Specifically, the total predictive variance is defined as:
\begin{align}\label{eq:total_prediction}
\sigma_{\mathrm{total},k}^{2^{(n_o)}}(\mathbf{s}_k)
&= \sigma_{\mathrm{aleatoric},k}^{2^{(n_o)}}(\mathbf{s}_k)
+ \sigma_{\mathrm{epistemic},k}^{2^{(n_o)}}(\mathbf{s}_k), \\[4pt]
\sigma_{\mathrm{aleatoric},k}^{2^{(n_o)}}(\mathbf{s}_k)
&= \frac{1}{H_{\mathrm{mc}}}
\sum_{\eta=1}^{H_{\mathrm{mc}}}
\sigma_{\eta}^{2^{(n_o)}}(\mathbf{s}_k), \\[4pt]
\sigma_{\mathrm{epistemic},k}^{2^{(n_o)}}(\mathbf{s}_k)
&= \frac{1}{H_{\mathrm{mc}}}
\sum_{\eta=1}^{H_{\mathrm{mc}}}
\bigl[\mu_{\eta}^{(n_o)}(\mathbf{s}_k)-\bar\mu^{(n_o)}(\mathbf{s}_k)\bigr]^{2}, \\[6pt]
\bar\mu^{(n_o)}(\mathbf{s}_k)
&= \frac{1}{H_{\mathrm{mc}}}
\sum_{\eta=1}^{H_{\mathrm{mc}}}
\mu_{\eta}^{(n_o)}(\mathbf{s}_k).
\end{align}
where \(\mu_{\eta}^{(n_o)}(\mathbf{s}_k)\) and \(\sigma_{\eta}^{2^{(n_o)}}(\mathbf{s}_k)\) denote the mean and variance outputs of the LSTM neural network on the \(\eta\)-th stochastic forward pass, respectively; \(H_{\mathrm{mc}}\) is the total number of Monte Carlo samples. The term \(\sigma_{\mathrm{aleatoric},k}^{2^{(n_o)}}\) captures irreducible, input-dependent aleatoric uncertainty in the prediction; \(\sigma_{\mathrm{epistemic},k}^{2^{(n_o)}}\) quantifies epistemic uncertainty by measuring the dispersion of the predicted means around their Monte Carlo average \(\bar\mu^{(n_o)}(\mathbf{s}_k)\);  \(\sigma_{\mathrm{total},k}^{2^{(n_o)}}(\mathbf{s}_k)\) is the total prediction uncertainty. 

Although $\sigma_{\mathrm{total},k}^{2^{(n_o)}}(\mathbf{s}_k)$ quantifies the total predictive variance in the output space $\mathbf{o}_k$, the ultimate objective is to express this uncertainty in the original space $\mathbf{u}_i$.  In the MLP–LSTM and MPNN–LSTM architectures, the mapping from \(\mathbf{u}_i\) to \(\mathbf{o}_k\) consists exclusively of affine transformations, so the predictive variance in the \(\mathbf{u}_i\) domain can be recovered exactly through standard linear uncertainty‐propagation rules. In contrast, the AE–LSTM model employs a nonlinear autoencoder to project \(\mathbf{u}_i\) into a latent space, precluding an analytic inversion for variance propagation. To recover uncertainty in the original input space, a reparameterization‐based Monte Carlo scheme is employed. After computing the total predictive variance \(\sigma_{\mathrm{total},k}^{2^{(n_o)}}(\mathbf{s}_k)\) and the predictive mean \(\bar{\mu}^{(n_o)}(\mathbf{s}_k)\), the stochastic variable \(\gamma^{(n_o)}_{k}(\mathbf{s}_k)\) is defined as:
\begin{equation}\label{reparameter}
\gamma^{(n_o)}_{k}(\mathbf{s}_k)
=\;\bar{\mu}^{(n_o)}(\mathbf{s}_k)
\;+\;\sigma_{\mathrm{total},k}^{(n_o)}(\mathbf{s}_k)\,\epsilon,\quad
\epsilon\sim\mathcal{N}(0,1)
\end{equation}
where $\epsilon$ is drawn once per Monte Carlo iteration and is independent of both the wavelet index $k$ and the output component $n_o$. By reusing the same $\epsilon$ across all indices for each sample, the reconstructed time series will be free of spurious zig–zag fluctuations and remain smooth. A total of $H_{mc,\gamma}$ samples of $\gamma^{(n_o)}_{k}(\mathbf{s}_k)$ are generated for each $k$ and $n_o$. Each sampled $\gamma^{(n_o)}_{k}$ is passed through the necessary reconstruction to its counterpart in the physical space of $\mathbf{u}_i$. The empirical variance of these reconstructions yields the total predictive uncertainty in $\mathbf{u}_i$ for each time step $i$ and component $n$, computed independently. For the present study, $H_{mc,\gamma} = H_{mc} = 50$ for simplicity.
The uncertainty estimations mentioned above are computed exclusively during the inference phase. During training, the network parameters are optimized as in a standard dropout-regularized model, and the loss function is specified by the negative log-likelihood loss as mentioned in Sec.~\ref{Sec:Aleatoric_Uncertainty}. Furthermore, the reparameterization-based Monte Carlo scheme described above is also employed for both the MLP–LSTM and MPNN–LSTM architectures to maintain consistency across all proposed approaches.

\section{Case Study}
\label{Sec: Case Study}

\subsection{Preamble}
\label{Sec:Preamble}

To demonstrate the effectiveness of the three proposed methodologies, two distinct case studies involving seismic stochastic excitations are presented.

The first case study considers a multi-degree-of-freedom (MDOF) shear-building structure modeled using the Bouc–Wen hysteretic formulation. This structure incorporates uncertainties in structural parameters such as building density, damping ratio, and Young's modulus. Comprising 40 degrees of freedom, the structural random variables are assumed to be perfectly correlated throughout the building cross-section. This simpler configuration allows the evaluation of proposed methods under conditions of relatively low output dimensionality and straightforward geometric configuration.

The second case study investigates a fiber-discretized frame with Giuffré-Menegotto-Pinto material model, while considering large displacement effects. This more complex structure includes uncertainties in damping, Young’s modulus, and yield strength. The structure has a total of 777 degrees of freedom, with the structural random variables assumed to be uncorrelated between the floor groups identified in Table \ref{tab:Bouc_SectionSize}. This scenario is selected specifically to evaluate the proposed methodologies in a realistic, high-dimensional structural system, providing insights into their performance under more demanding and practical conditions.

A site-specific seismic hazard characterization corresponding to a 10\% probability of exceedance in 50 years for subsurface conditions classified as Site Class D was adopted \citep{molina2016seismic,ASCE_2022}. This hazard model served as the basis for generating stochastic ground‐motion records in accordance with the methodology of \citet{rezaeian2010}. Each record comprised 30 s of motion, which was subsequently extended by appending 30 s of zero acceleration to permit the structure’s free vibration. Consequently, the structural response was simulated over a total duration of 60 s, using a time increment of 0.005 s to yield 12000 discrete time steps.

\subsection{Case Study 1: MDOF Bouc–Wen Shear Building}
\label{Sec:Case Study 1}

\subsubsection{Setup}
\label{Sec:Case1_Setup}

\begin{figure}[t]
\centering
\includegraphics[width=0.55\linewidth]{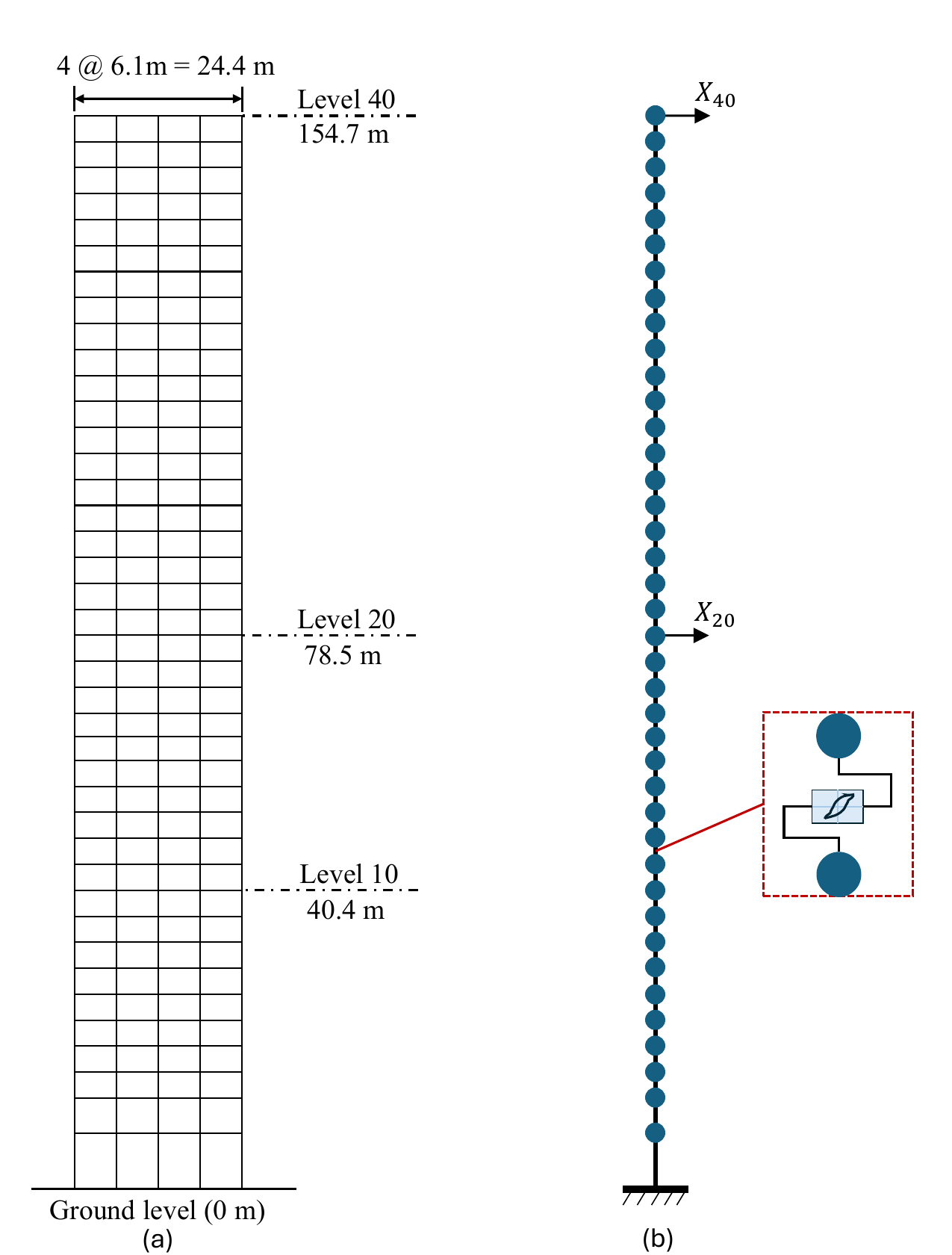}
\caption{2D steel frame with Bouc–Wen model of hysteresis : (a) structural layout; and (b) shear building model.}
\label{bouc_layout}
\end{figure}

The first case study, illustrated in Fig.~\ref{bouc_layout}, examines a two-dimensional steel frame modeled with Bouc–Wen hysteretic nonlinearity and subjected to stochastic seismic excitation characteristic of downtown San Francisco. The geometry and mass distribution follow the design of \citet{molina2016seismic}, yielding a total height of 154.7 m, with a 6.1 m ground story and upper stories each 3.9 m tall. At every level, four bays of 6.1 m span are arranged in series to produce an overall plan width of 24.4 m. Structural mass is lumped at the floor levels. Gravity and lateral loads are resisted by AISC wide-flange steel beams and square hollow-section columns, whose sectional properties are summarized in Table.~\ref{tab:Bouc_SectionSize}. Under a shear-building assumption, the lateral response is governed by nonlinear interstory restoring forces simulated through the Bouc–Wen model. The structural system is developed in MATLAB and subsequently solved with the Runge-Kutta method. For this case study, the uncertainties considered in the building mass density, damping, and Young’s modulus are summarized in Table~\ref{tab:Bouc_parameter}, together with their corresponding probability distributions.

The training, validation, and testing datasets consisted of 1500, 200, and 200 independent realizations of both structural parameters and stochastic ground‐motion inputs, respectively. To isolate the effect of structural‐parameter variability, an additional set of 400 “deterministic” samples was generated by fixing all structural properties at their mean values while using the same excitation time histories as in the validation and test sets. The size of the training set was chosen by varying the number of samples until the MLP–LSTM baseline achieved an average mean‐squared error on the validation data that was approximately one order of magnitude lower than the MSE of the “deterministic” samples on the validation dataset. Prior to model training, the excitation and response time series were normalized by their expected peak values and subjected to a sixth‐order Daubechies wavelet transform with scale parameter 3. Structural‐property features were standardized by subtracting the training sample mean and dividing by the standard deviation.

%
A broad range of configurations were evaluated on the proposed MLP–LSTM baseline, and the set of parameters yielding the lowest average validation error was subsequently adopted for both the MPNN–LSTM and AE–LSTM approaches. For the AE–LSTM, the latent-space dimension $n_r$ was fixed at 3, a choice guided by an empirical procedure in which $n_r$ was incrementally increased until further dimensions produced only marginal improvements in reconstruction error. This strategy for selecting the autoencoder’s latent dimensionality mirrors that employed by Simpson et al. \citep{simpson2021machine} and Zhang et al. \citep{zhang2024rlstm}. 

\begin{table*}[t]
  \centering
  \begin{threeparttable}
    \caption{Section sizes used in the Bouc-Wen shear building.}
    \label{tab:Bouc_SectionSize}
    \footnotesize
    \setlength{\tabcolsep}{20pt}
    \begin{tabular}{l c c c}
      \hline
      Floors  & Beams           & Interior columns [cm] & Exterior columns [cm] \\ \hline
      1       & W36$\times$282  & 66$\times$7.6          & 66$\times$6.4          \\
      2-10   & W36$\times$282  & 56$\times$7.6          & 51$\times$6.4          \\
      11-20  & W36$\times$194  & 51$\times$5.0          & 51$\times$5.0          \\
      21-30  & W33$\times$169  & 46$\times$2.5          & 46$\times$2.5          \\
      31-40  & W27$\times$84   & 46$\times$1.9          & 46$\times$1.9          \\ \hline
    \end{tabular}
    \begin{tablenotes}
      \footnotesize
      \item Note: Columns sizes: (outer side size) $\times$ (wall thickness) in cm.
    \end{tablenotes}
  \end{threeparttable}
\end{table*}
\begin{table*}[ht]
  \centering
  \footnotesize
  \setlength{\tabcolsep}{12pt} 
  \caption{Statistical properties of uncertain structural parameters}
  \label{tab:Bouc_parameter}
  \begin{tabular}{l c c c}
    \toprule
    Parameter         & Mean            & COV  & Distribution \\
    \midrule
    Young’s Modulus   & 200 GPa   & 0.04 & Lognormal       \\
    Building Density    & 150 $kg/m^3$  & 0.1 & Normal        \\
    Damping Ratio     & 0.005 & 0.40 & Lognormal    \\
    \bottomrule
  \end{tabular}
\end{table*}

\subsubsection{Results}
\label{Sec:Case1_Results}

To evaluate the predictive accuracy of the proposed models on the test dataset, three normalized error metrics were employed: the mean squared error (MSE), root mean squared error (RMSE), and mean absolute error (MAE). Denoting the number of spatial degrees of freedom by \(n = 40\), the number of test realizations by \(N_s = 200\), and the length of each time‐history by \(T\), the normalized MSE is defined as:
\begin{equation}\label{eq:MSE}
\mathrm{MSE}
= \frac{1}{n\,N_s\,T}
  \sum_{j=1}^{N_s}
  \sum_{i=1}^{T}
  \bigl\|
    (\mathbf{u}_{ij} - \hat{\mathbf{u}}_{ij})
    \oslash \bar{\mathbf{u}}_{\mathrm{peak}}
  \bigr\|_{2}^{2},
\end{equation}
where \(\mathbf{u}_{ij}\in\mathbb{R}^n\) denotes the true displacement vector for sample \(j\) at time step \(i\), \(\hat{\mathbf{u}}_{ij}\) is the corresponding model prediction, and \(\bar{\mathbf{u}}_{\mathrm{peak}}\) represents the component‐wise average peak displacement computed over the training set. The RMSE and MAE are given by:
\begin{equation}\label{eq:RMSE}
\mathrm{RMSE}
= \frac{1}{N_s}
  \sum_{j=1}^{N_s}
  \sqrt{
    \frac{1}{n\,T}
    \sum_{i=1}^{T}
    \bigl\|
      (\mathbf{u}_{ij} - \hat{\mathbf{u}}_{ij})
      \oslash \bar{\mathbf{u}}_{\mathrm{peak}}
    \bigr\|_{2}^{2}
  },
\end{equation}
\begin{equation}\label{eq:MAE}
\mathrm{MAE}
= \frac{1}{n\,N_s\,T}
  \sum_{j=1}^{N_s}
  \sum_{i=1}^{T}
  \bigl\|
    (\mathbf{u}_{ij} - \hat{\mathbf{u}}_{ij})
    \oslash \bar{\mathbf{u}}_{\mathrm{peak}}
  \bigr\|_{1},
\end{equation}
where \(\|\cdot\|_1\) denotes the element‐wise \(L^1\) norm.

Table~\ref{boucwen_results} summarizes the MSE, RMSE, and MAE for the Bouc–Wen case study. All three approaches achieve similarly low error levels, with the MLP–LSTM model attaining the most favorable performance across all metrics. Because these global metrics yield only aggregate measures, a sample‐ and component‐wise MSE is introduced to provide insight into model behavior on individual degrees of freedom:
\begin{equation}\label{eq:MSE_sam}
\mathrm{MSE}_{\alpha}^{(j)}
= \frac{1}{n_{\alpha}\,T}
  \sum_{i=1}^{T}
  \bigl\|
    (\mathbf{u}_{ij}^{(\alpha)} - \hat{\mathbf{u}}_{ij}^{(\alpha)})
    \oslash \bar{\mathbf{u}}_{\mathrm{peak}}^{(\alpha)}
  \bigr\|_{2}^{2},
  \quad \alpha \in \{X, Z, R\},
\end{equation}
where \(\mathbf{u}_{ij}^{(\alpha)}\) and \(\hat{\mathbf{u}}_{ij}^{(\alpha)}\) denote the true and predicted displacement components in direction \(\alpha\) (horizontal \(X\), vertical \(Z\), or rotational \(R\)), \(n_{\alpha}\) is the number of degrees of freedom in direction \(\alpha\), and \(\bar{\mathbf{u}}_{\mathrm{peak}}^{(\alpha)}\) is the corresponding average peak response over the training set. Under the shear‐building assumption adopted for the Bouc–Wen structure, only the horizontal components (\(\alpha = X\)) are nonzero; accordingly, \(\mathrm{MSE}_{X}^{(j)}\) fully characterizes sample‐wise performance in this case.

\begin{figure}[]
\centering
\includegraphics[width =0.5\linewidth]{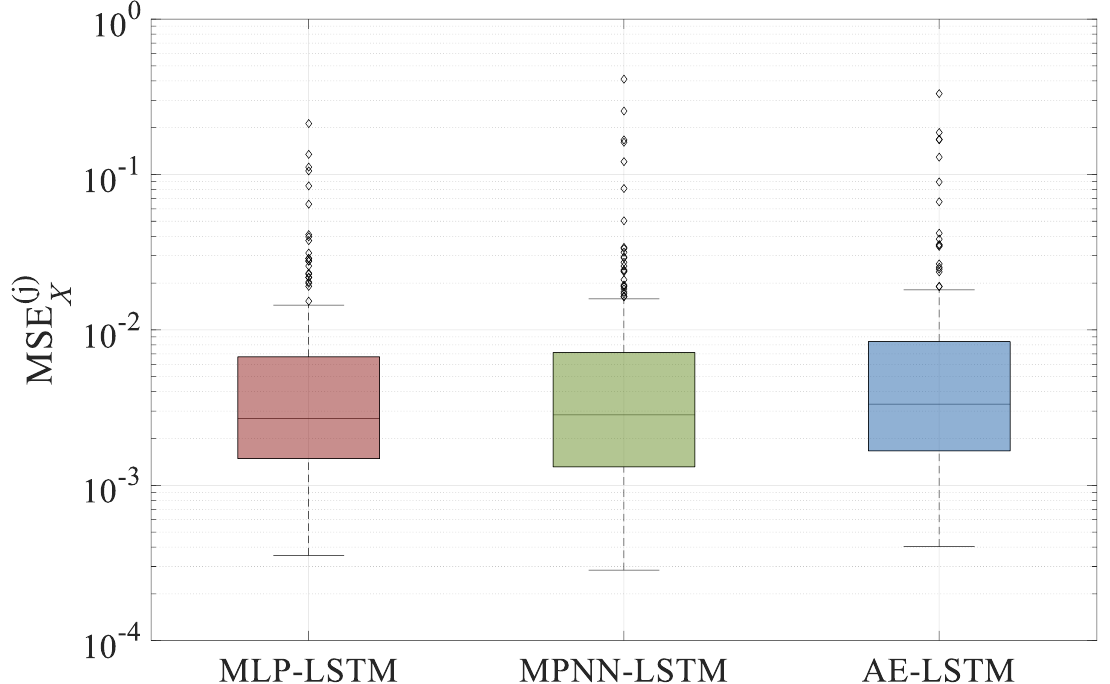}
\caption{Box plot of the sample-wise X-direction displacement response MSE of different proposed schemes for the Bouc-Wen case study.}
\label{Bouc_Wen_Box}
\end{figure}

Fig.~\ref{Bouc_Wen_Box} present the box‐plot distributions for each proposed method, revealing that while all three approaches produce similar overall distributions, the MLP–LSTM consistently achieves a more concentrated error profile with lower median and interquartile range. This performance advantage arises from the simplicity of the Bouc–Wen case study: the shear‐building model features a straightforward, line‐connected mass‐element configuration and a low system dimensionality, which limits the benefit of message‐passing operations in the MPNN–LSTM. Likewise, the modest dimensionality diminishes the value of latent‐space compression in the AE–LSTM. As a result, the AE–LSTM incurs the largest errors of the three methods, since its total error comprises both the autoencoder’s reconstruction error and the LSTM’s prediction error, which together exceed the direct mapping error of the MLP–LSTM.

\begin{figure}[]
\centering
\includegraphics[scale=1,width=\textwidth]{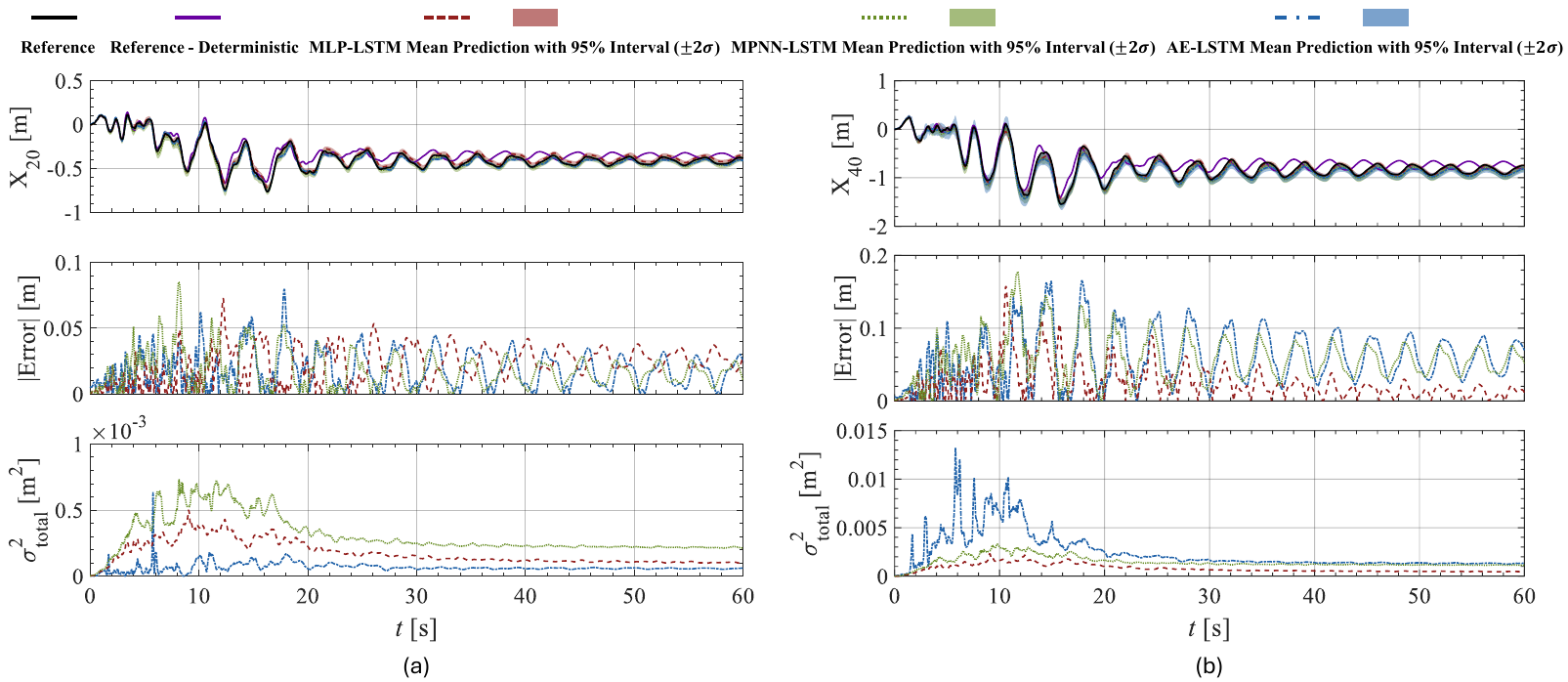}
\caption{Sample displacement time histories for Bouc-Wen case study: (a) horizontal displacement of node 20; (b) horizontal displacement of node 40.}
\label{Bouc_Wen_time_history}
\end{figure}

\begin{figure}[]
\centering
\includegraphics[width =1\linewidth]{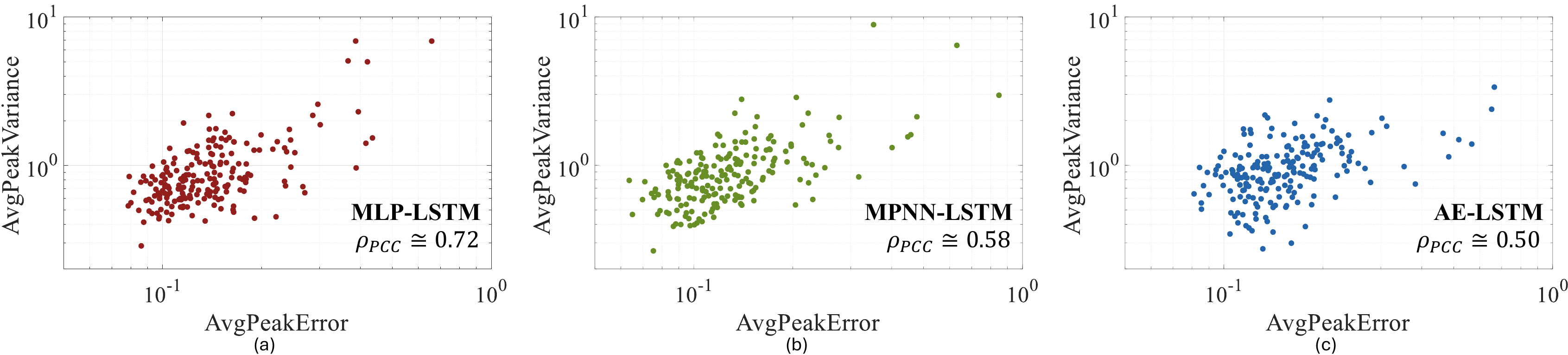}
\caption{Correlation between maximum MAE with maximum variance over test sample: (a) MLP-LSTM scheme; (b) MPNN-LSTM scheme; (c) AE-LSTM. Note: the figures are in log-log scale for easier visualization.}
\label{Bouc_Corr}
\end{figure}

Figs.~\ref{Bouc_Wen_time_history}(a) and \ref{Bouc_Wen_time_history}(b) present horizontal displacement time histories, absolute errors, and predictive variances for nodes 20 and 40 in Fig.~\ref{bouc_layout}. Time histories exhibiting approximately median absolute errors were chosen for illustration. The curve labeled “Reference-Deterministic” corresponds to the ground-truth simulation with all structural parameters fixed at their mean values, whereas “Reference” denotes the full stochastic simulation. The discrepancy between these two reference traces underscores the effect of parameter variability on the system’s dynamic evolution. In Fig. \ref{Bouc_Wen_time_history}(a), all proposed models achieve comparable accuracy levels, while in Fig.~\ref{Bouc_Wen_time_history}(b), the MLP–LSTM model attains slightly lower absolute errors. Furthermore, the ground truth values are covered in the 95\% prediction interval for most of the time for all of the methods. Across all methods, the predicted variance peaks in regions of peak error, demonstrating that the variance provides a meaningful measure of each model’s uncertainty. Therefore, it is of interest to determine and validate whether the peak error values and peak variance values have any correlations across the entire test samples and other DOFs; thus, Fig.~\ref{Bouc_Corr} presents the correlation between the average peak absolute error (AvgPeakError) and average peak predictive variance (AvgPeakVariance). AvgPeakError is computed as:
\begin{equation}\label{eq:avg_max_error}
\mathrm{AvgPeakError}_{j} =
\frac{1}{n}
\sum_{d=1}^{n}
\frac{
\max\limits_{1 \leq i \leq T}
\left| \mathbf{u}_{ijd} - \hat{\mathbf{u}}_{ijd} \right|
}{
\bar{\mathbf{u}}_{\mathrm{peak},d}
}
\end{equation}
where \( \left| \mathbf{u}_{ijd} - \hat{\mathbf{u}}_{ijd} \right| \) denotes the absolute error between the predicted and reference displacement at time step \(i\), for test sample \(j\) and degree of freedom (DOF) \(d\), as illustrated in the $|{\rm Error}|$ plots of Fig.~\ref{Bouc_Wen_time_history}. The operator \( \max_{1 \leq i \leq T}\) extracts the maximum absolute error over the entire time history for each DOF (i.e., the peak value of the corresponding $|{\rm Error}|$ plot). To mitigate the disproportionate influence of DOFs with inherently larger displacement amplitudes during the averaging process, each peak error is normalized by the expected peak displacement \( \bar{u}_{\mathrm{peak},d} \). The resulting metric thus represents the mean normalized peak error across all DOFs for a given test realization.

An analogous formulation is used to compute the average peak predictive variance (AvgPeakVariance), wherein the absolute error term \( \left| \mathbf{u}_{ijd} - \hat{\mathbf{u}}_{ijd} \right| \) is replaced by the total predictive variance \( \sigma^{2}_{\text{total}} \), and normalization is performed using the expected peak predictive variance \( \bar{\sigma}_{\mathrm{peak},d} \). The Pearson correlation coefficient ($\rho_{PCC}$) is used for analyzing the correlation between average peak predictive variance and average peak absolute error.

Figure \ref{Bouc_Corr} demonstrates a moderate positive correlation between the average maximum predictive variance and the average peak absolute error across all methods. This trend indicates that, for a given sample, larger peak errors tend to coincide with higher predictive uncertainty. Although predictive variance does not serve as a perfect indicator for error magnitude, it nonetheless provides a valuable criterion for applications such as active‐learning schemes: by prioritizing samples with elevated predictive variance, one can identify those realizations most likely to exhibit poor prediction accuracy without resorting to costly ground‐truth simulations.
%
\begin{table*}[!t]
\begin{center}
\footnotesize
\setlength{\tabcolsep}{20pt}
\caption{Summary of the performance of the various proposed schemes for the Bouc-Wen case study.}
\label{boucwen_results}
\begin{tabular}{lccc}
\hline
\multicolumn{1}{l}{Proposed Schemes}      & \multicolumn{1}{c}{MAE}   & \multicolumn{1}{c}{MSE}    & \multicolumn{1}{c}{RMSE}  \\ 
\hline
MLP–LSTM   & \textbf{0.0616} & \textbf{0.0090} & \textbf{0.0680} \\
MPNN–LSTM  & 0.0642          & 0.0115          & 0.0700          \\
AE–LSTM    & 0.0674          & 0.0114          & 0.0741          \\
\hline
\end{tabular}
\vspace{1ex}

\footnotesize{Note: Boldface highlights the best performer for each metric.}
\end{center}
\end{table*}
%
%
%
%
%
%
%
%
%
%
%
%
%
\subsection{Case Study 2: Fiber-Discrtized Nonlinear Frame}
\label{Sec:Case Study 2}
\label{Sec:Case2_Setup}
\subsubsection{Setup}
\label{Sec:Case2_Setup_sub}
\begin{figure}[t]
\centering
\includegraphics[width=0.47\linewidth]{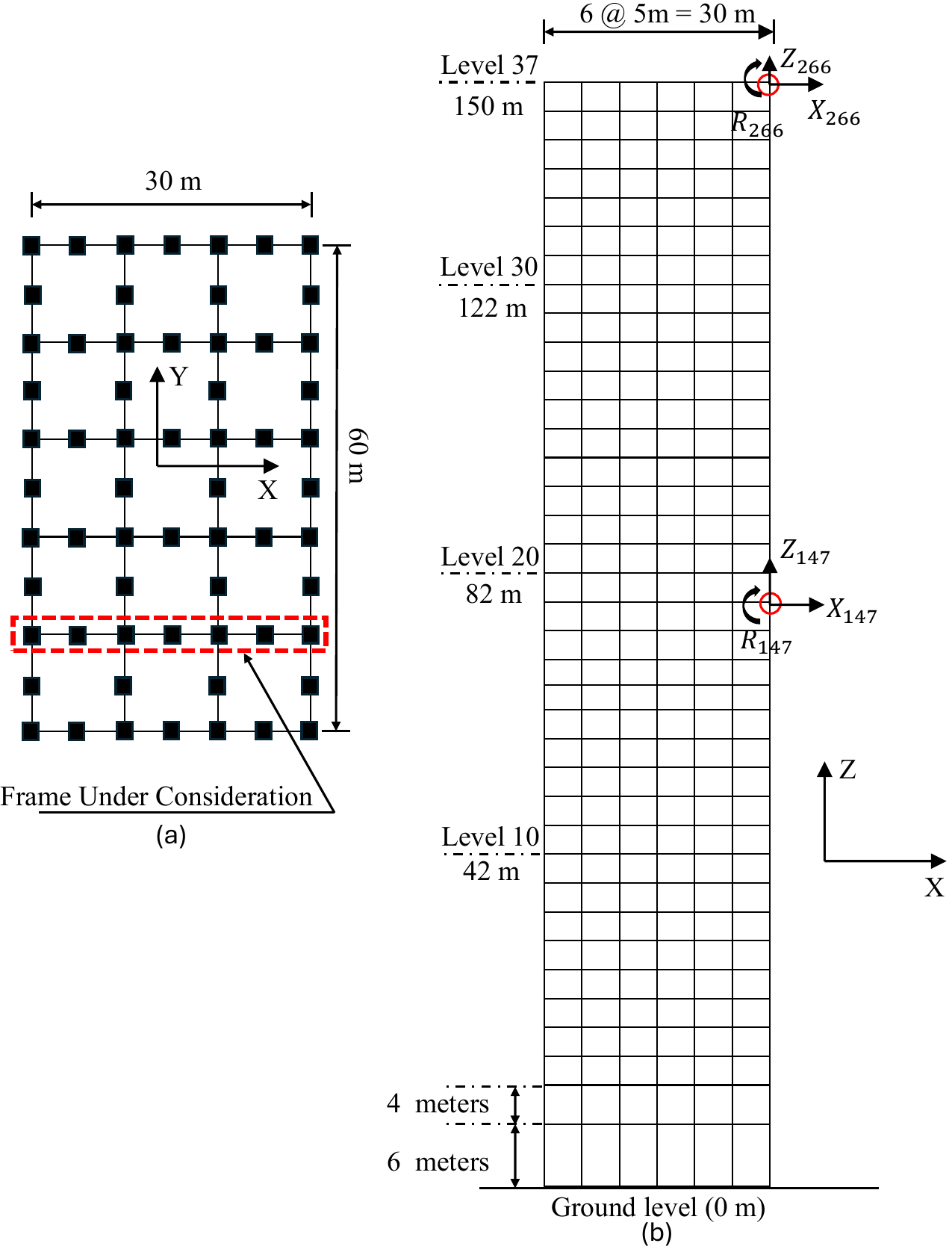}
\caption{Layout of the 2D 37-story six-bay steel frame: (a) Plan view; and (b) Elevation view.}
\label{37_layout}
\end{figure}

The second case study examines a thirty‐seven–story, six‐span steel frame discretized with nonlinear fiber‐based finite elements, as depicted in Fig.~\ref{37_layout}. Geometric and sectional properties follow the designs of \citet{molina2016seismic} and \citet{chuang2017performance}. Columns are modeled as square hollow sections, whereas beams employ AISC W24 wide‐flange profiles; their dimensions are reported in Table.~\ref{tab:SectionSize}. Floor masses combine member self‐weight with a uniform carried mass based on a building density of 100 kg/m$^3$. The entire structural system was implemented in OpenSees \cite{opensees} using displacement‐based fiber elements with five Gauss–Legendre integration points per member length. Each fiber follows an Giuffré-Menegotto-Pinto material model, and large‐displacement effects are accounted for using corotational transformation. Uncertainties in damping ratio, Young’s modulus, and yield strength were introduced according to the probabilistic distributions summarized in Table.~\ref{tab:37Story_parameter_stats}.

Data for model training, validation, and testing consist of 2000, 200, and 200 independent realizations of both structural properties and stochastic ground‐motion records, respectively. The procedures for sample‐size selection and data normalization replicate those of the first case study. Similar to the first case study, a sixth‐order Daubechies wavelet transform with scale parameter three was used. 
%
%
Following the same criterion used previously, the latent‐space dimension $n_{r}$ was determined to be 40 for all employed architectures.

\begin{table}[b]
\begin{center}
\begin{threeparttable}
\caption{Section sizes used in the steel frame.}
\footnotesize
\setlength{\tabcolsep}{20pt}	
\label{tab:SectionSize}
\begin{tabular}{l c c}
\hline
Floors & Beams & Box columns [cm] \\ \hline
1 - 10 &  W24$\times$192 & 51.25$\times$2.5 \\
11 - 20 &  W24$\times$192 &  51.25$\times$2.5 \\
21- 30 &  W24$\times$103 & 41$\times$2.0 \\
31 - 37 &  W24$\times$103 &  35.9$\times$1.8 \\ \hline
\end{tabular}%
\begin{tablenotes}
\footnotesize
\item Note: Box column size defined as (width) $\times$ (thickness).
\end{tablenotes}
\end{threeparttable}
\end{center}
\label{Material}
\end{table}%
\begin{table*}[ht]
  \centering
  \footnotesize
  \setlength{\tabcolsep}{12pt} 
  \caption{Statistical properties of uncertainty structural parameters}
  \label{tab:37Story_parameter_stats}
  \begin{tabular}{l c c c}
    \toprule
    Parameter         & Mean            & COV  & Distribution \\
    \midrule
    Young’s Modulus   & 200 GPa   & 0.04 & Lognormal       \\
    Yield Strength    & 0.38 GPa  & 0.06 & Lognormal        \\
    Damping Ratio     & 0.015 & 0.40 & Lognormal    \\
    \bottomrule
  \end{tabular}
\end{table*}

\subsubsection{Results}
\label{Sec:Case2_Results}

\begin{figure}[]
\centering
\includegraphics[width =1\linewidth]{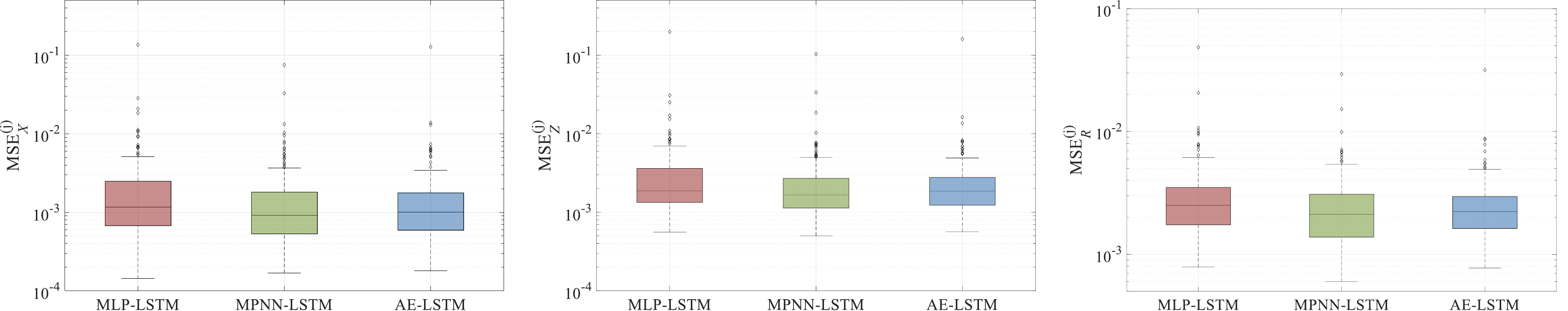}
\caption{Box plot of the sample-wise direction-wise displacement response MSE of different proposed schemes for the fiber-discretized nonlinear frame case study: (a) horizontal direction; (b) vertical direction; (c) rotational direction.}
\label{37_box}
\end{figure}

In the fiber-discretized nonlinear frame case study, performance was quantified using the mean squared error (MSE), root mean squared error (RMSE), and mean absolute error (MAE) metrics, ensuring consistency with the Bouc–Wen analysis. Table \ref{37story_results} demonstrates that all proposed architectures yielded comparably low error magnitudes; however, unlike the Bouc–Wen case, the MPNN–LSTM and AE–LSTM models significantly outperformed the MLP–LSTM. To mitigate the influence of outliers, box-plot distributions in Fig.~\ref{37_box} reveal that both the MPNN–LSTM and AE–LSTM exhibit lower errors across all response components. The enhanced geometric complexity of the fiber-discretized frame—characterized by beams and columns with heterogeneous properties—facilitates more effective message passing and richer graph representations in the MPNN–LSTM, thereby improving its accuracy relative to the MLP–LSTM. Moreover, the high dimensionality of this system renders the latent-space compression inherent to the AE–LSTM particularly advantageous, as it enables the LSTM to focus on temporal prediction within a reduced-dimensional space. Between the two advanced architectures, the MPNN–LSTM achieves the lowest errors, a result that may be attributed to reconstruction errors introduced by the autoencoder in the AE–LSTM.

\begin{figure}[]
\centering
\includegraphics[width =1\linewidth]{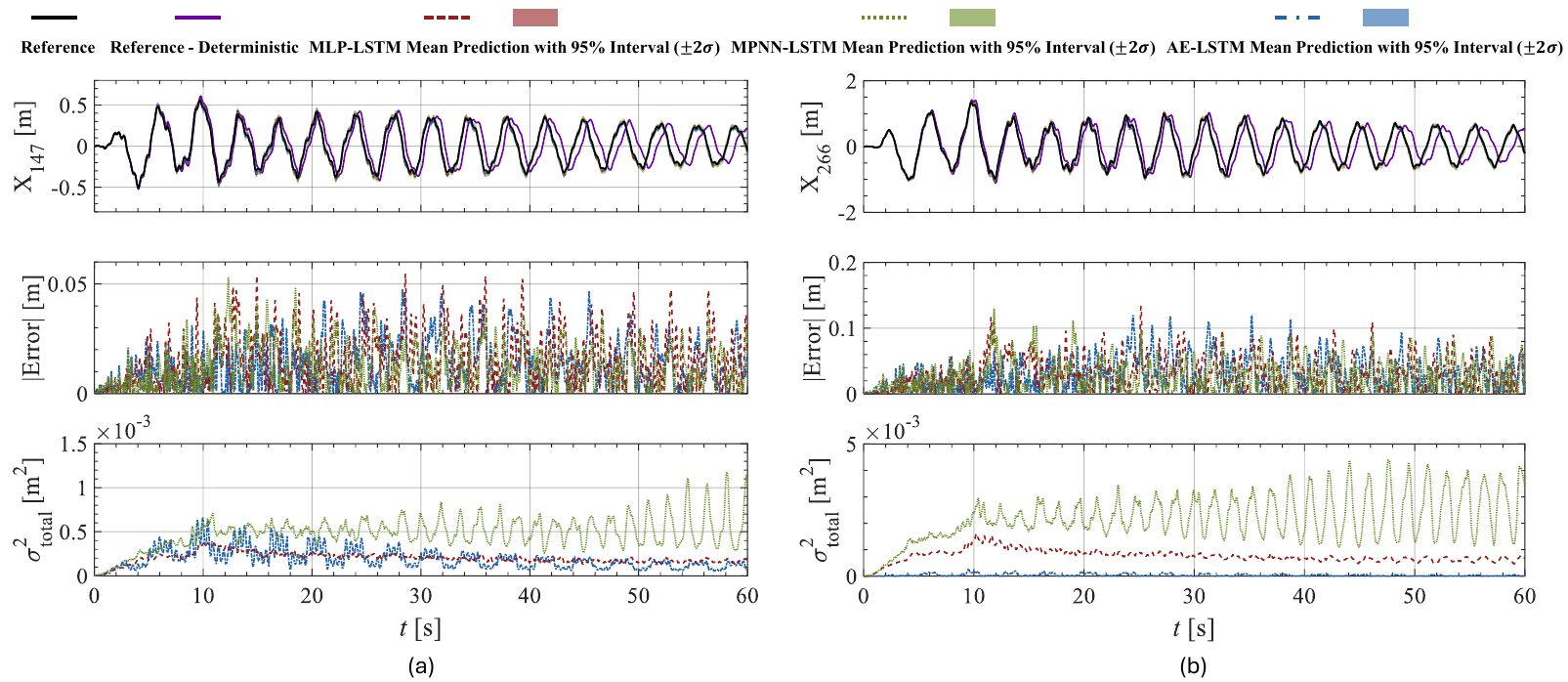}
\caption{Sample displacement time histories for fiber-discretized nonlinear frame case study: (a) horizontal displacement of node 147; (b) horizontal displacement of node 266.}
\label{37_time_historyX}
\end{figure}

\begin{figure}[]
\centering
\includegraphics[width =1\linewidth]{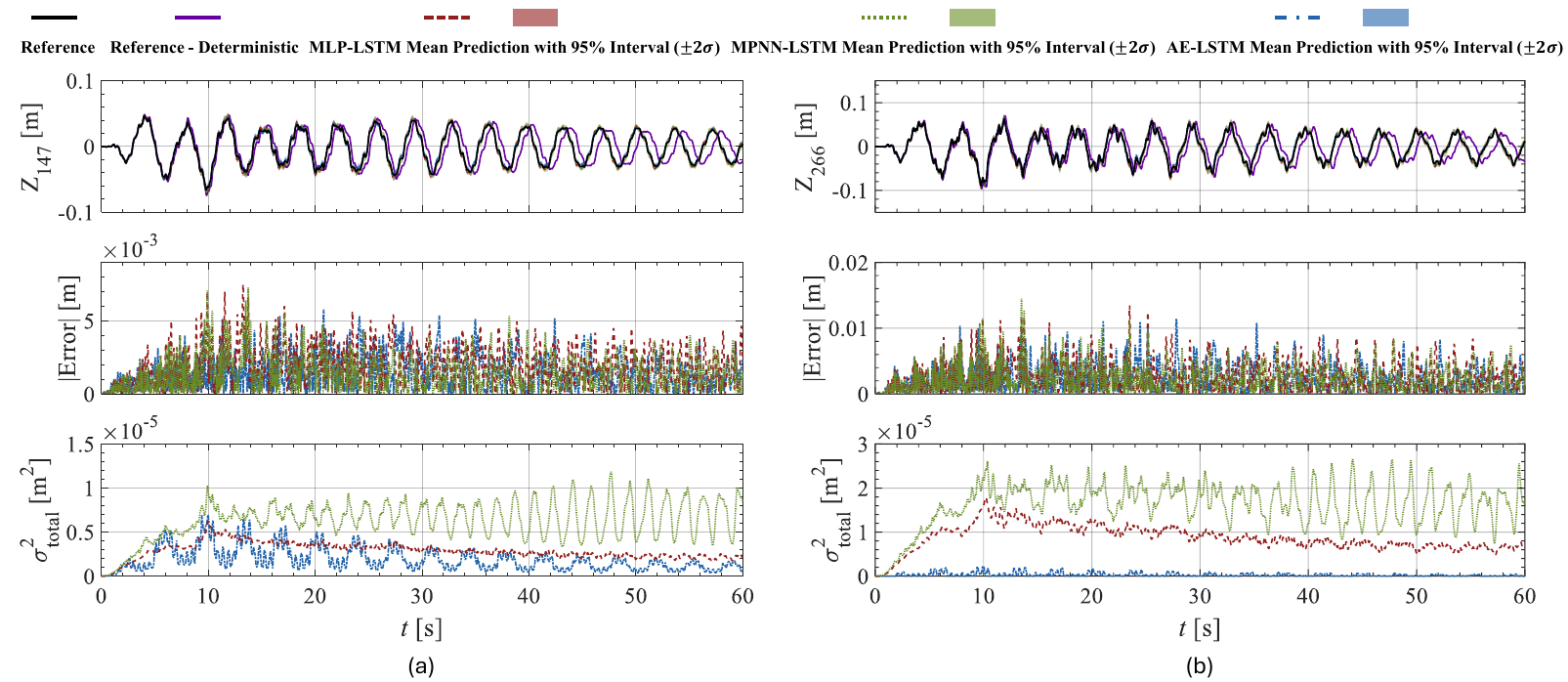}
\caption{Sample displacement time histories for fiber-discretized nonlinear frame case study: (a) vertical displacement of node 147; (b) vertical displacement of node 266.}
\label{37_time_historyZ}
\end{figure}

\begin{figure}[]
\centering
\includegraphics[width =1\linewidth]{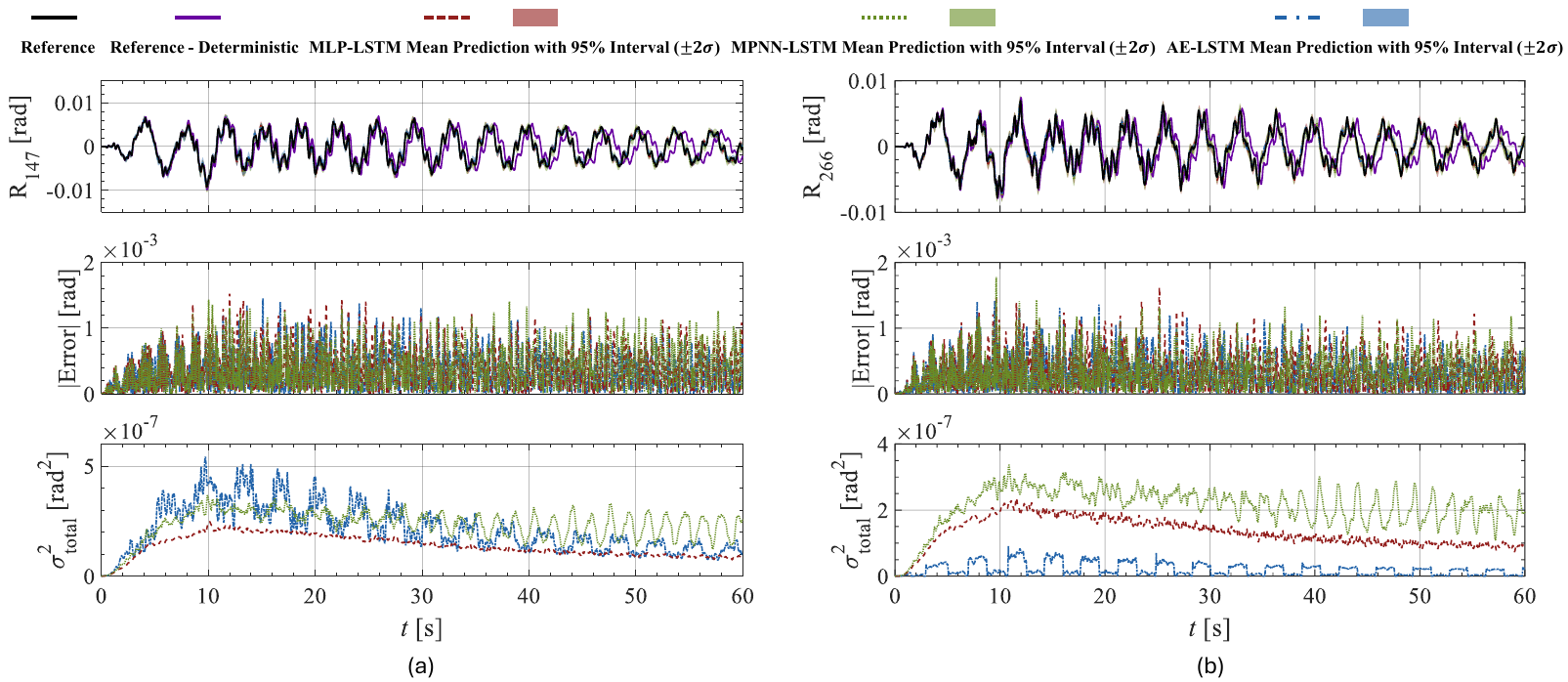}
\caption{Sample displacement time histories for fiber-discretized nonlinear frame case study: (a) rotational displacement of node 147; (b) rotational displacement of node 266.}
\label{37_time_historyR}
\end{figure}

Figs.~\ref{37_time_historyX}, \ref{37_time_historyZ}, and \ref{37_time_historyR} present representative time-history predictions alongside the corresponding absolute errors and predictive variances for the horizontal, vertical, and rotational displacements at nodes 147 and 266, respectively.  The curve labeled “Reference-Deterministic” corresponds to the simulation with all structural parameters fixed at their mean values, whereas “Reference” denotes the full stochastic simulation. The divergence between these two reference traces underscores the influence of parameter variability on the system’s dynamic evolution. All proposed models closely reproduce the stochastic ground truth, maintaining low absolute errors throughout the time history for different components across the building.  Moreover, relative to the Bouc–Wen case study, the predictive variance observed in the fiber-discretized nonlinear frame is substantially lower, which may be attributed to the superior prediction accuracy achieved by the proposed methods in this case study. Notably, the AE–LSTM architecture exhibits particularly low variance at node 266, which may be attributed to the autoencoder’s latent representation being less sensitive to minor input perturbations (i.e., the decoder’s Jacobian norm with respect to the input is small at this node). 

\begin{figure}[]
\centering
\includegraphics[width =1\linewidth]{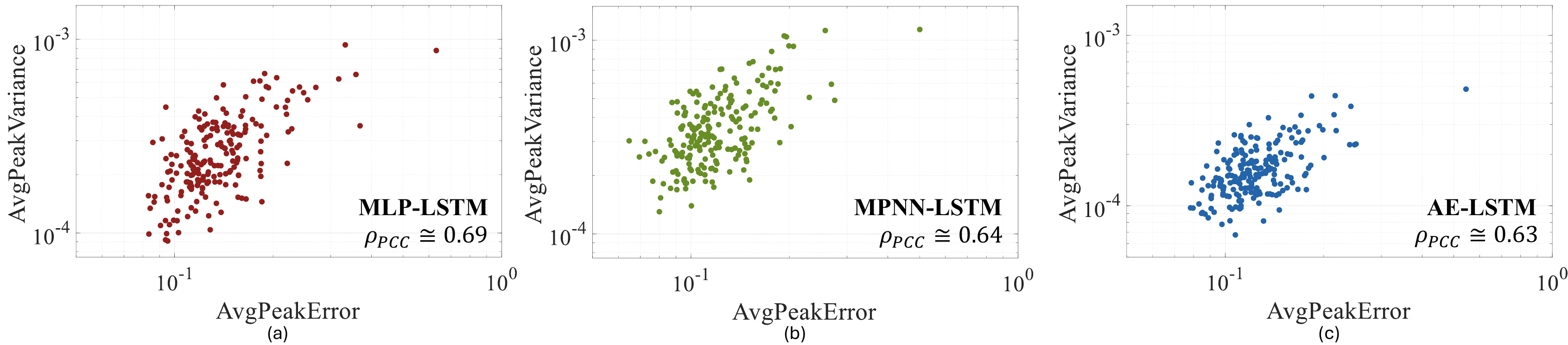}
\caption{Correlation between maximum MAE with maximum variance over test sample: (a) MLP-LSTM scheme; (b) MPNN-LSTM scheme; (c) AE-LSTM. Note: the figures are in log-log scale for easier visualization.}
\label{37_Corr}
\end{figure}

Analogous to the Bouc–Wen case study, Fig.~\ref{37_Corr} examines the relationship between the average peak absolute error and the average peak predictive variance for the fiber-discretized frame. A moderate positive correlation is observed across all models, supporting the conclusion that higher predictive uncertainty tends to coincide with greater prediction error. The consistency of this trend across both case studies affirms the generality of the observed relationship, indicating that the association between predictive variance and absolute error is not coincidental but instead reflects a broader characteristic of the proposed modeling framework.
%
%
\begin{table*}[!t]
\begin{center}
\footnotesize
\setlength{\tabcolsep}{20pt}
\caption{Summary of the performance of the various proposed schemes for the fiber-discretized nonlinear frame case study.}
\label{37story_results}
\begin{tabular}{lccc}
\hline
\multicolumn{1}{l}{Proposed Schemes}      & \multicolumn{1}{c}{MAE}   & \multicolumn{1}{c}{MSE}    & \multicolumn{1}{c}{RMSE}  \\ 
\hline
MLP–LSTM   & 0.0356          & 0.0034           & 0.0444 \\
MPNN–LSTM  & \textbf{0.0314}          & \textbf{0.0025}          & \textbf{0.0391}          \\
AE–LSTM    & 0.0324          & 0.0027          & 0.0404          \\
\hline
\end{tabular}
\vspace{1ex}

\footnotesize{Note: Boldface highlights the best performer for each metric.}
\end{center}
\end{table*}
%
%
%
%
%
%
%
%
%
%
%
%

\newpage

\section{Summary and Conclusions}
\label{Sec:Conclusion}

This study proposes three distinct long short-term memory (LSTM)–based metamodeling frameworks—MLP–LSTM, MPNN–LSTM, and AE–LSTM—for capturing the nonlinear dynamic response histories of structural systems characterized by both structural parameter uncertainty and stochastic excitation, while simultaneously quantifying predictive uncertainty. The efficacy of the proposed approaches is demonstrated through two representative case studies: a relatively simple multi-degree-of-freedom Bouc–Wen system and a more complex, nonlinear, fiber-discretized steel moment-resisting frame. Both systems are subjected to stochastic seismic loading and incorporate uncertainty in their structural parameters.

Across both case studies, all three metamodeling approaches yield low prediction errors, affirming their suitability for modeling complex structural responses under uncertainty. In the Bouc–Wen case study, the MLP–LSTM model achieves the highest predictive accuracy among the proposed methods. However, in the fiber-discretized steel frame case study—which involves significantly greater structural and geometric complexity—the MPNN–LSTM and AE–LSTM models outperform the MLP–LSTM, highlighting the advantage of their enhanced representational capacity in more demanding settings.

These findings suggest that while the MLP–LSTM is well-suited for simpler problems, the MPNN–LSTM and AE–LSTM architectures are more appropriate for high-dimensional or structurally intricate systems, where their capacity to capture spatial dependencies or compress latent representations becomes advantageous. Additionally, a consistent correlation is observed between predictive variance and prediction error across both case studies. Although not exact, this correlation offers a valuable indicator for gauging model confidence, and it holds particular promise for future applications such as active learning, wherein predictive uncertainty may inform adaptive sampling strategies and improve training efficiency.


\section*{Data Availability}
All codes will made available on GitHub, \url{https://github.com/hatila822/Deep_Learning_with_Parametric_and_Prediction_Uncertainty}, upon publication of the paper.

\section*{Acknowledgments}
This research effort was supported in part by the National Science Foundation (NSF) under Grant No. CMMI-2118488. This support is gratefully acknowledged.








%
\appendix
%
%

\bibliographystyle{elsarticle-num-names}

\bibliography{Bibliography}







\end{document}